\documentclass[letterpaper, 10 pt, conference]{ieeeconf}  

\IEEEoverridecommandlockouts                              

\usepackage{graphics} 
\usepackage{epsfig} 

\makeatletter
\let\NAT@parse\undefined
\makeatother

\usepackage{color}
\usepackage{url}
\usepackage{hyperref}
\usepackage{amsmath}
\usepackage{relsize}

\usepackage{color}
\usepackage{url}
\usepackage{hyperref}
\usepackage{amsmath}
\usepackage{relsize}
\usepackage{graphicx}
\usepackage{amssymb}
\usepackage{booktabs}
\usepackage{multirow}
\usepackage{rotating}
\usepackage{float}
\usepackage{soul}

\DeclareMathOperator*{\argmax}{arg\,max}

\usepackage{color}
\definecolor{jeremy_color}{rgb}{0,.7,.7}
\definecolor{patrick_color}{rgb}{.6,.4,.05}
\definecolor{charlie_color}{rgb}{0,0,0.8}
\definecolor{cody_color}{rgb}{0.75,0.25,0.0}
\definecolor{youliang_color}{rgb}{0.35,0.75,0.0}
\definecolor{revision_color}{rgb}{0,0,0}

\newcommand{\comment}[1]{} 

\newcommand{\jeremy}[1]{\textcolor{jeremy_color}{\comment{jeremy: #1}}}

\newcommand{\ck}[1]{\textcolor{charlie_color}{\comment{charlie: #1}}}

\newcommand{\youliang}[1]{\textcolor{youliang_color}{\comment{youliang: #1}}}

\newcommand{\method}[0]{ForceSight}
\newcommand{\network}[0]{ForceSight}

\title{ForceSight: Text-Guided Mobile Manipulation with Visual-Force Goals}

\author{Jeremy A. Collins$^{1\textbf{*}}$, Cody Houff$^{1\textbf{*}}$, You Liang Tan$^{1\textbf{*}}$, Charles C. Kemp$^{1}$
\\
$^\textbf{*}$Equal contribution
}

\newenvironment{first_caption}
  {\par\footnotesize}
  {\par\addvspace{\bigskipamount}}

\begin{document}

\twocolumn[{%
\renewcommand\twocolumn[1][]{#1}%
\maketitle
\thispagestyle{empty}
\pagestyle{empty}

\begin{center}
\centering
\vspace{-8mm}
\includegraphics[width=0.95\linewidth]{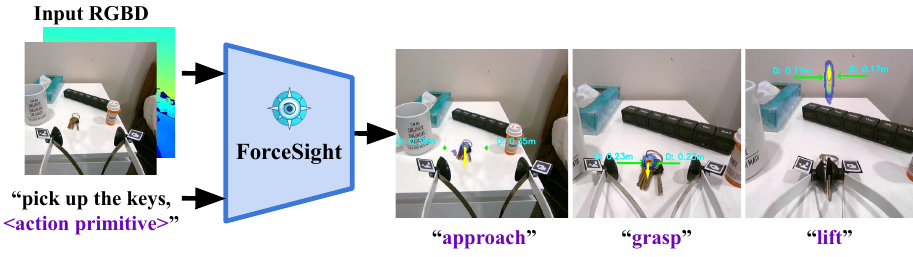}
\end{center}
\begin{first_caption}
 Fig. 1. \network{} is an RGBD-adapted, text-conditioned vision transformer. Given an RGBD image and a text prompt, ForceSight produces visual-force goals for a mobile manipulator. Action primitives, shown below each image, are appended to the text input by a simple low-level controller. $D$ refers to the estimated depth of the goal location in meters.
\end{first_caption}
\vspace{-3mm}
}]

\setcounter{figure}{1}  
\setcounter{footnote}{1}


\footnotetext{The authors are with the Institute for Robotics and Intelligent Machines at the Georgia Institute of Technology (GT). This work was supported in part by NSF Award \# 2024444 and AI-CARING Award \# 2112633. Charles C. Kemp contributed to this work as an associate professor at GT. He now works full-time for Hello Robot Inc., which sells the Stretch RE1.}

\begin{abstract}
We present ForceSight, a system for text-guided mobile manipulation that predicts visual-force goals using a deep neural network. Given a single RGBD image combined with a text prompt, ForceSight determines a target end-effector pose in the camera frame (kinematic goal) and the associated forces (force goal). Together, these two components form a visual-force goal. Prior work has demonstrated that deep models outputting human-interpretable kinematic goals can enable dexterous manipulation by real robots. Forces are critical to manipulation, yet have typically been relegated to lower-level execution in these systems. When deployed on a mobile manipulator equipped with an eye-in-hand RGBD camera, ForceSight performed tasks such as precision grasps, drawer opening, and object handovers with an 81\% success rate in unseen environments with object instances that differed significantly from the training data. In a separate experiment, relying exclusively on visual servoing and ignoring force goals dropped the success rate from 90\% to 45\%, demonstrating that force goals can significantly enhance performance. The appendix, videos, code, and trained models are available at \href{https://force-sight.github.io/}{https://force-sight.github.io/}.

\end{abstract}

\section{Introduction}

Robotic manipulation has significantly benefited from the integration of tactile and force information. These modalities enable direct perception and control of contact with the environment, which can be advantageous. For example, grasping a small or flat object off a surface can benefit from first making fingertip contact with the surface and then sliding the fingertips across the surface to pick up the object \cite{babin2018picking}. Success depends on fingertip force that is high enough to maintain contact with the surface and grasp the object, but low enough to slide the fingertips. In general, grip force and applied force provide strong cues that appropriate contact has been achieved for a task. 




We introduce \method{}, a transformer-based, text-conditioned robotic planner that outputs visual-force goals, enabling the human-interpretable execution of tasks in novel environments with unseen object instances. \method{} uses an RGBD-adapted, text-conditioned vision transformer to encode an RGBD image from a gripper-mounted camera and output a visual-force goal relevant to the current subtask. A visual-force goal consists of a kinematic goal and a force goal. The kinematic goal specifies a target configuration for the end-effector as a 3D position, a yaw angle, and the distance between the fingertips. The force goal specifies a target grip force and a target applied force measured by a wrist-mounted force-torque sensor. 

We present the following contributions:
\begin{itemize}
    \item \textbf{Visual-Force Planner:} We present a model that infers visual-force goals given an RGBD image from an eye-in-hand camera and a natural language prompt and show that force goals significantly improve performance over visual servoing alone. 
    \item \textbf{Task Execution System:} We present a system that uses visual-force goals from our model to perform a variety of tasks with unseen object instances in novel environments. 
    \item \textbf{Open Source:} We release our code, dataset, and trained models.
\end{itemize}

\section{Related Work}
\label{sec:citations}

\begin{figure*}
  \centering
  \includegraphics[width=0.97\linewidth]{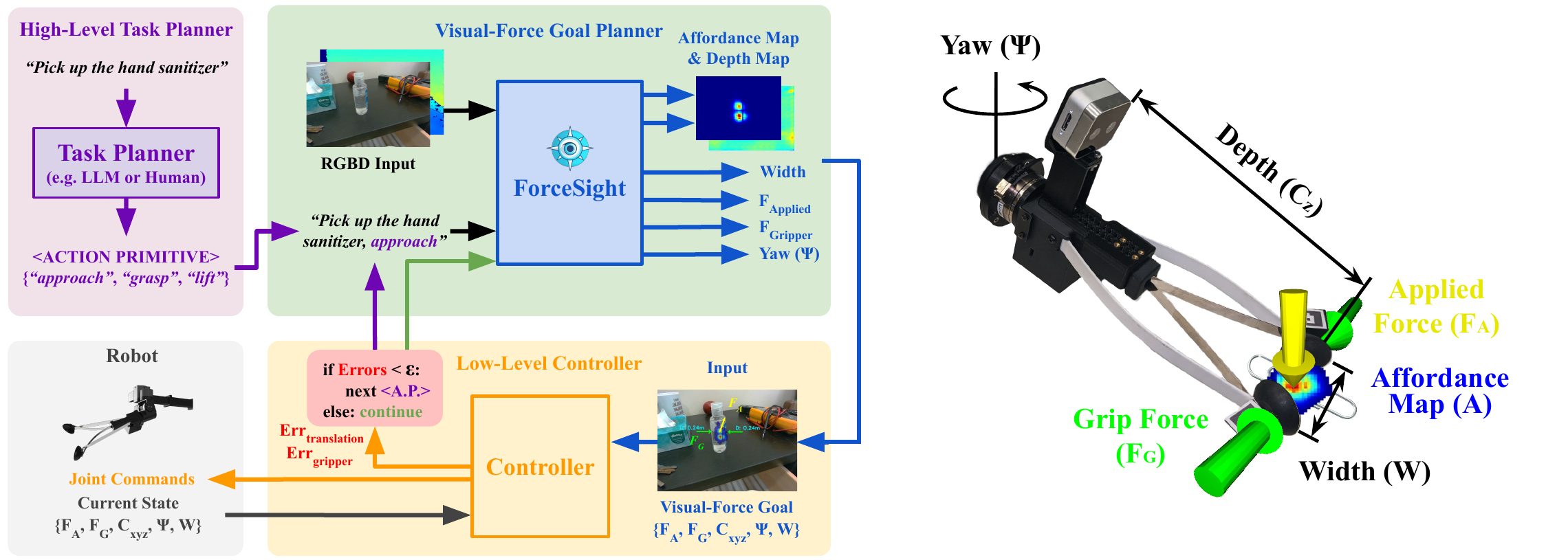}
  \caption{\label{appendix:sys-archi} \textbf{Left:} Overall System Architecture. \label{fig:visual-force goals} \textbf{Right:} A visual-force goal includes target fingertip locations (green arrow tips), target grip force (green arrow magnitudes), and target applied force (yellow arrow). The affordance map represents a probability distribution for the gripper position goal in pixel space. The goals are represented with respect to the camera frame.}
  \label{fig:sys-archi-diagram}
   \vspace{-5mm}
\end{figure*}

Imitation learning has been widely adopted in robotics, enabling robots to learn from human demonstrations for complex tasks. Recent work has explored imitation learning to create robot policies from combined text and image input \cite{brohan2022rt, shridhar2022cliport, shridhar2023perceiver, jang2022bc, brohan2023rt}. Videos have also shown to be an effective source of data for learning policies, affordances, and value functions associated with human behavior \cite{bahl2023affordances,bahl2022human, ma2023vip, nair2022r3m, wang2023mimicplay, mandikal2022dexvip}. However, practical considerations such as occlusion and depth ambiguity can limit the performance of vision-based systems. Our system uses force as a complementary modality that provides direct measurements of mechanical interaction. Prior work has used kinematic objectives for robotic planning \cite{shridhar2022cliport, shridhar2023perceiver, bahl2023affordances, kulkarni2019unsupervised, manuelli2022kpam, james2022coarse}. We show that using force and kinematic objectives in tandem can improve performance in robotic tasks.

While some methods \cite{shridhar2022cliport, shridhar2023perceiver, valassakis2022demonstrate, johns2021coarse} learn behaviors in a data-efficient manner, and generalize to object pose via data augmentation or explicit object representations, they often come at the cost of being restricted to narrow environments with fixed cameras. Other works prioritize versatility but demand an extensive amount of data \cite{brohan2022rt, jang2022bc}. Our approach lies between these two regimes, generalizing across environments and perspectives with a modest amount of data collection effort.





Research suggests force goals can be predicted from vision. Everyday mechanisms involve predictable forces that can be inferred \cite{jain2013improving}. Humans quickly learn to predict appropriate forces for precision lifting \cite{flanagan2006control}. Classes of objects have predictable masses, and evidence indicates that humans can visually estimate the weight of novel objects \cite{peters2015smaller}. 

Force sensing can improve robustness and generalization. For example, forces remain consistent in the presence of widely varying visual phenomena. And, mechanisms have predictable forces that can be captured by a person or a robot and used by a distinct robot, demonstrating the potential for learned force goals to generalize across robots \cite{jain2013improving}.



The idea of combined visual-force servoing has been explored with both classical and data-driven methods for tasks such as peg-in-hole insertion and contour following \cite{baeten2003integrated, almaghout2021robotic}. The conceptualization of course-to-fine visual servoing has also been proposed in prior work using data-driven methods, where many initial poses lead to a singular ``bottleneck pose" \cite{valassakis2022demonstrate, johns2021coarse, lu2022cfvs}. However, these methods were demonstrated to work for single-step tasks with seen object instances and controlled environments. Our method performs sequences of action primitives by chaining together visual-force goals that are similar to bottleneck poses. In contrast to methods with similar data collection schemes, \method{} is capable of composing behaviors to complete multi-step tasks in a variety of environments and generalizes to objects semantically similar to those in the training set.


\begin{figure*}
  \centering
  \includegraphics[width=0.8\linewidth]{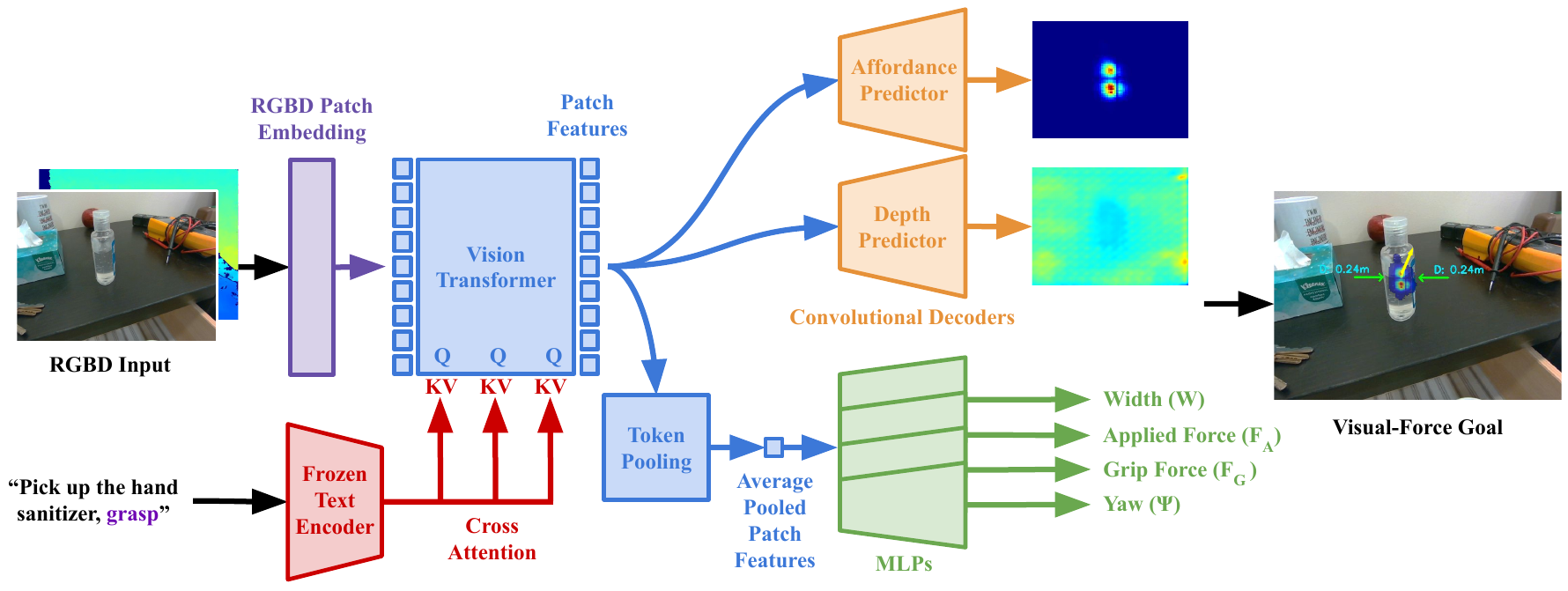}
  \caption{\network{} is a text-conditioned RGBD vision transformer. An RGBD image is first divided into patches and passed into an RGBD-adapted patch encoder that transforms image patches into image tokens. These image tokens are fed into a vision transformer. After every transformer block inside the vision transformer, the visual features are conditioned on a text embedding via cross-attention to produce text-conditioned image patch features. These patch features are passed into two convolutional decoders to produce an affordance map and a depth map. The patch features are additionally average pooled and passed into several MLPs in order to predict the gripper width, applied force, grip force, and yaw.}
  \label{fig:architecture}
   \vspace{-6mm}
\end{figure*}

\section{\method{}: A Force-Based Robotic Planner}
\label{sec:method}

We present \method{}, a system that represents robotic tasks as sequences of visual-force goals. A visual-force goal includes target fingertip locations in the camera frame, a target grip force, and a target force applied by the gripper. 


\subsection{Representing Visual-Force Goals}

We define a visual-force goal $G$ as the combination of a kinematic goal $G_K$ and a force goal $G_F$. (Figure \ref{fig:visual-force goals}) The kinematic goal, which defines the desired 4-DOF pose (x, y, z, yaw) of the gripper, is parameterized as $G_K = \{ C_{xy}, C_z, \psi, W \}$.


$C_{xy} \in \mathbb{R}^2$ is the 2D pixel coordinate of the 3D goal position projected onto the input image, and $C_{z} \in \mathbb{R}$ is the depth estimate of the 3D goal position in the camera frame. Together, $C_{xy}$ and $C_{z}$ define a 3D point corresponding to the desired gripper location. In order to obtain a prediction for $C_{xy}$, we introduce an affordance map  $\mathcal{A}\in \mathbb{R}^{H \times W}$, which is a probability distribution in pixel space describing the likelihood that the gripper should move to each pixel location. $C_{xy}$ is obtained by taking $\argmax_{x,y} \mathcal{A}$, the most likely pixel coordinate associated with the goal position of the gripper. We additionally introduce a depth map $D\in \mathbb{R}^{H \times W}$. The depth estimate $C_z$  is obtained by taking the pixel value from the 2D depth map at $C_{xy}$.  $W$ (gripper width) is the Euclidean distance between the predicted fingertip locations, and $\psi$ is the predicted yaw of the gripper with respect to the camera frame, in radians. From these values, we can derive the 4-DOF kinematic goal of the gripper, i.e. the desired gripper position, yaw, and aperture.

In addition to a kinematic goal, we define a force goal $G_F = \{F_{A}, F_G\}$. The applied force $F_A \in \mathbb{R}^{3}$ is a vector that represents the desired applied force associated with the next step of a task, measured by a force/torque sensor mounted to the wrist of the robot, then transformed into the camera frame. Grip force, $F_G$, is a scalar value representing the magnitude of the compressive force between the fingertips. $F_G$ is estimated with a small neural network (see Section \ref{sec:data-collection} for details).

Given an RGBD observation and a text prompt, \method{} predicts visual-force goals $G_K$ and $G_F$ one keyframe into the future, with respect to the current camera frame. The output representations for \method{} and eye-in-hand camera setup make the future predictions more amenable to visual servoing \cite{hutchinson1996tutorial}, which uses closed-loop control for improved robustness. Moreover, these predictions could be useful to other embodied systems since the predicted goals are not dependent on the camera mounting location (Figure \ref{fig:All_tasks}). Consequently, a low-level policy can be adapted to visually servo a robot to reach its predicted goals.

\subsection{Data Collection}
\label{sec:data-collection}
We collect a dataset $D$, with $D_i = \{I,T,C_{LR},F_G, F_A\}$, where $I \in \mathbb{R}^{H \times W \times 4}$ is an RGBD image captured by a gripper-mounted camera, $T$ is a text prompt associated with a task and includes an action primitive, $C_{LR} \in \mathbb{R}^{2 \times 3}$ is a set of two 3D fingertip locations in the camera frame associated with the next keyframe, $F_G$ is the grip force, and $F_A$ is the applied force. These combined elements constitute the data points in the dataset, providing a rich representation of a robot's interaction with its environment. From these datapoints, the kinematic and force goals ($G_K$ and $G_F$) that serve as ground truth for our network can be derived.


This dataset contains over 26,000 high-quality datapoints, the equivalent of roughly 10,000 task demonstrations. Data was collected by teleoperating a Stretch RE1 mobile manipulator \cite{kemp2022design} for approximately 30 hours. We accomplish this rate of approximately 11 seconds per demonstration by associating multiple input images with the same visual-force goal, essentially collecting multiple views of a given goal from various perspectives. We also sampled from states that are unlikely to occur during a successful execution in order to promote recovery from errors. This allows our algorithm to be resilient to potentially imprecise predictions, thereby strengthening the robustness of the system. This is related to the concepts of funneling \cite{mason1985mechanics} and pre-image backchaining \cite{lozano1984automatic} that have inspired recent work in robust feedback motion planning \cite{majumdar2017funnel}.

To further enhance the efficiency of data collection, objects relevant to other tasks were deliberately included in many input images. This allows us to map the same input image to other goals that are visible in the image, consequently generating substantially more data points per image in the dataset.


In order to collect visual data, we mount an Intel® RealSense™ D405 \cite{intelrealsenseDepthCamera} to the robot's gripper to capture the RGBD image $I$. The force applied to the gripper is measured by a wrist-mounted force/torque sensor \cite{ATI}. To obtain the grip force, which is not natively available on the Stretch RE1 gripper, we train an MLP to estimate the grip force $F_G$, given the gripper motor position, motor current, and fingertip positions. To provide ground truth for the grip force model, we grasp a force/torque sensor \cite{ATI} at various grip strengths and grasp widths, and record the measured force magnitude.

To collect ground truth, the gripper's fingertips are first localized in the image via ArUco tags attached to the gripper, and a transformation is applied to map these to the point at the center of each fingertip's surface, which we denote as the fingertip locations $C_{LR} \in \mathbb{R}^{2 \times 3}$. The fiducial markers could potentially be replaced with other methods of pose estimation. Subsequently, the robot's forward kinematics are utilized to map the 3D fingertip locations from the camera frame where the goal was achieved to the camera frame where the input image was captured.

\subsection{Network Architecture}
\vspace{-5pt}



Our proposed architecture leverages a large-scale Vision Transformer \cite{dosovitskiy2020image} (ViT-large, 304M parameters) and a frozen T5 text encoder \cite{raffel2020exploring} to output precise visual-force goals (Figure \ref{fig:architecture}). Visual-force goals are composed of fingertip locations, grip force, and applied force, enabling robotic manipulation based on tactile objectives.

We initialize our vision transformer with weights from pre-training on ImageNet 21k \cite{ridnik2021imagenet}. We introduce an enhanced patch embedding layer that accepts RGBD inputs to this pre-trained network. To accommodate depth alongside RGB channels, we add a fourth input channel to the patch embedding projection, much like the early fusion technique described in \cite{tziafasearly}. We initialize the weights of this additional channel with the average of the existing RGB channel weights, enabling smooth integration of depth information without forgetting information learned during pre-training.

The text-conditioned component of \network{} relies on a frozen pre-trained T5 text encoder, which generates a text embedding vector that offers context for a given task. To effectively utilize this text information, we incorporate a cross-attention mechanism across all layers of the vision transformer, allowing the network to represent relationships between text prompts and visual features at multiple levels of abstraction. Inspired by developments in text-conditioned image generation \cite{rombach2022high}, this operation applies cross-attention using projections of visual features as query vectors and projections of text features as key and value vectors.



The network output includes a $224\times224$ affordance map $\mathcal{A}$ and a depth map $D$. The ViT image encoder generates patch features associated with the input image's particular regions, which are transformed into an affordance map and depth map through a convolutional decoder. We apply a weighted cross-entropy loss to the affordance map, translating the task into a pixel-wise classification problem. The depth map prediction is supervised via an $L_1$ loss, masked by the ground truth affordance map, to focus only on the predicted gripper position. Hence, the depth map is conditioned on the affordance map, producing a depth estimate $C_z$ in pixel space. Here, we define $C_z = D[C_{xy}]$, where $C_{xy}$ represents the coordinate with the maximum likelihood in the affordance map, given by $C_{xy} = \argmax_{x, y}(\mathcal{A})$. This representation allows our network to propose multiple hypotheses, exhibiting robustness in scenarios where multiple visible objects are semantically relevant to the task.

Finally, the patch features are average-pooled and fed into several multi-layer perceptrons (MLPs). These MLPs estimate task execution parameters such as gripper width, yaw, applied force, and grip force. 




\subsection{System Architecture}
\label{sys-arch}
Our low-level controller receives kinematic and force goals located in the camera frame from \network{} and executes incremental joint commands after each image frame observation in order to achieve these goals (Figure \ref{fig:sys-archi-diagram}). Using an eye-in-hand camera, the low-level controller uses visual-force servoing to move the gripper closer to the visual-force goal. This approach reduces relative error and is insensitive to global calibration.


The objective of the low-level controller is to jointly minimize the kinematic and force errors. To better incorporate both kinematic and force modalities into the movement error,  we define a joint objective combining both errors. This is expressed as a movement command  $M_\text{end-effector} \in \mathbb{R}^3$ and $M_\text{gripper} \in \mathbb{R}$ in Cartesian space, and is then executed by the controller in a step-wise manner.
\begin{equation}
\begin{split}
M_{\text{end-effector}} &= K_{ee} ( T_k \mathcal{E}_{\text{translation}} + \lambda_\text{applied} T_f \mathcal{E}_{\text{applied-force}} )\\
M_{\text{gripper}} &= K_{g} ( \mathcal{E}_{\text{width}} + \lambda_\text{grip} \mathcal{E}_{\text{grip-force}} )\\
\end{split}
\end{equation}

Where $K_{ee}$ and $K_{g}$ are proportional gain matrices, $T_k$ and $T_f$ are matrices that transform kinematic and force errors to robot motions, $\mathcal{E}$ denotes errors relative to the goals, and $\lambda$ denotes weights on force errors relative to kinematic errors.

\subsection{Action Primitives}

The use of Large Language Models (LLMs) has demonstrated its efficacy in addressing sequential long-horizon robotic tasks \cite{lin2023text2motion, vemprala2023chatgpt, ahn2022can, driess2023palm, huang2022inner}.  In language-based tasks, it is often possible to identify shared subgoals across different tasks. By leveraging positive transfer of action subgoals among tasks, it becomes feasible to generalize primitive actions for robot tasks. To facilitate the transition between subgoals,  we incorporate action primitives such as \textit{approach, grasp, ungrasp, lift, and pull}, which are appended to the model prompt input. The low-level controller switches to the next action primitive once the errors between current states and target goals are adequately minimized, as outlined in the overall system architecture in Figure \ref{fig:sys-archi-diagram}. This can result in the system appearing to make multiple attempts prior to achieving success, serving as a form of error recovery. We implement the switching between action primitives using predetermined sequences incremented by a simple state machine, but one could plausibly automate this process with the use of an LLM, as is demonstrated on our website. \jeremy{add GPT-4 screenshot to website}

\section{Training Details}

For training, we processed 224x224 RGBD images. Ground truths for the affordance map take the form of multi-hot encodings, using circles with a radius of 10 pixels instead of single pixels to indicate the tool center point's coordinates. This denser representation proved efficient in generating rich heatmaps without compromising performance.

To improve model robustness and encourage generalization, the RGB channels of our input images underwent brightness, saturation, contrast, and hue augmentations. Additionally, during preprocessing, we applied a data filtering step to exclude examples with ground truths lying outside the camera's field of view, ensuring minimal prediction inaccuracies due to field of view constraints.

\network{} was trained over 20 epochs, equating to 500,000 iterations, using batches of eight using the Adam optimizer \cite{kingma2014adam} with a learning rate of 5e-5.

The loss function for \network{} is given by:
\begin{align}
L = \sum_{i \in \mathcal{L}} \lambda_i L_i
\label{eq:L}
\end{align}
Where:
$\mathcal{L} = {A, D, F_A, F_G, W, \psi}$ represents the set of all loss components and $L_A$, $L_D$, $L_{F_A}$, $L_{F_G}$, $L_W$, and $L_{\psi}$ correspond to the affordance map weighted cross-entropy loss, masked depth map MAE, applied force MSE, grip force MSE, gripper width MSE, and yaw MSE respectively. Depth loss is masked by the affordance map ground truth, and thus is only applied at locations where the affordance map ground truth is nonzero.

\section{Experimental Results}
\label{sec:result}

\begin{table}
\small
\scriptsize
\centering
\begin{tabular}{c|c}
    \textbf{Task} & \textbf{Action Primitives}\\\hline
    Pick up the apple & Approach, Grasp, Lift  \\
    Pick up the medicine bottle & Approach, Grasp, Lift\\
    Pick up the keys & Approach, Grasp, Lift\\
    Pick up the paperclip & Approach, Grasp, Lift\\
    Pick up the hand sanitizer & Approach, Grasp, Lift \\
    Pick up the cup & Approach, Grasp, Lift\\
    Place object in the trash & Approach, Ungrasp\\
    Place object in the hand & Approach, Ungrasp\\
    Turn off the light switch & Approach, Push \\
    Open the drawer & Approach, Grasp, Pull\\
\end{tabular}
\caption{Tasks and action primitives.}
\label{tab:test-definition}
\vspace{-5mm}
\end{table}

We use a Stretch RE1 \cite{kemp2022design} from Hello Robot to conduct real-world experiments. To evaluate our model, we conduct experiments in held-out environments, including a mock bedroom and a real kitchen (Table \ref{tab:test-definition}). The robot exclusively interacts with unseen object instances, i.e. objects that are semantically similar to those seen during training but are visually distinct from objects in the training dataset.

\begin{figure}
  \centering
  \includegraphics[width=1\linewidth]{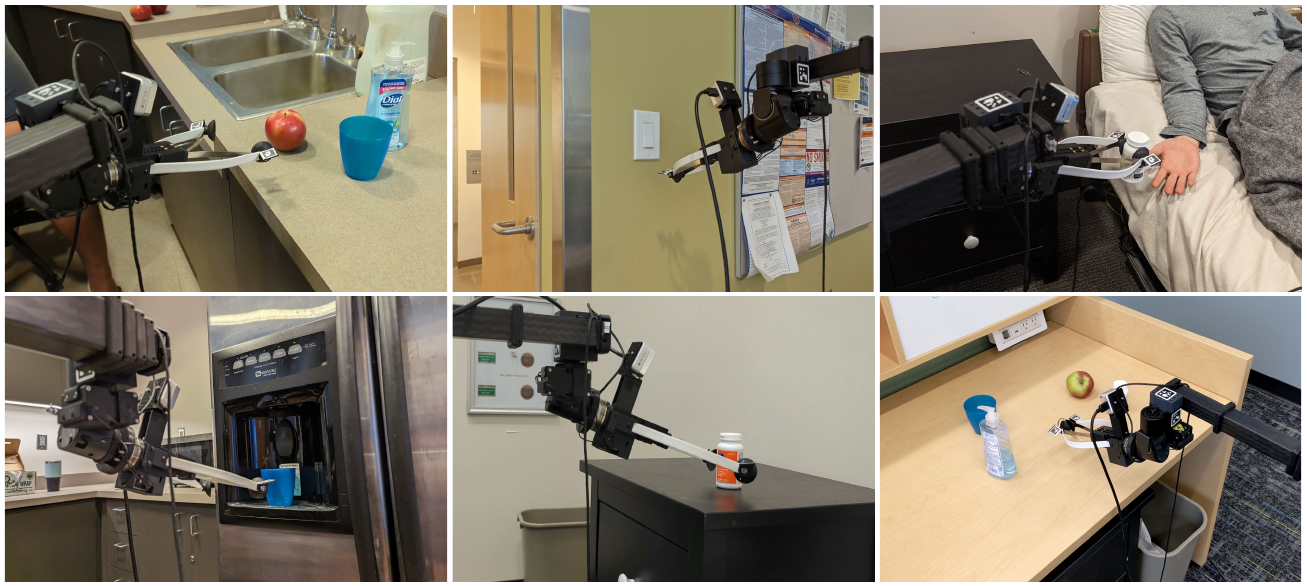}
  \vspace{-5mm}
  \caption{The real-world experiments consist of 6 picking tasks, 2 placing tasks, a drawer opening task, and a light switch flipping task, encompassing various objects and environments not present in the training set.}
  \label{fig:experiment-envs}
   \vspace{-5mm}
\end{figure}

\subsection{Experimental Setup}

To evaluate \method{}, we select a set of 10 household tasks and perform 10 trials on each, totaling 100 trials. We also perform real-world experiments on 5 ablations of our model, each for 20 trials. For each trial, we randomize the robot's pose within 3 unseen environments such that the target pose is in the camera's field of view and is within a distance of 1 meter. We also randomize the pose of the held-out target objects and distractor objects.
\ck{The following two sentences indicate that success is only based on the internal observations made by the robot, but clearly the experimenter is making decisions about success beyond this, such as whether the drawer was actually opened or the object was actually lifted rather than relying on what the robot thinks it's doing}
We define task success as achieving all subgoals associated with a task.






\subsection{Metrics}

We employ the following set of evaluation metrics:

\textbf{Average Fingertip Distance:} The mean $L_2$ distance between predicted and actual fingertip locations in meters (m). Smaller values signify better accuracy.

\textbf{RMSE:} Calculates the root-mean-square error between predicted and actual values of applied and grip forces ($F_A$ and $F_G$), reflecting force estimation accuracy.

\textbf{Task Success Rate:} The percentage (\%) indicating the system's success rate in completing tasks across trials.

\subsection{Baselines}

We selected Perceiver-Actor (PerAct) \cite{shridhar2023perceiver} as a baseline, training and evaluating it on our dataset. To make PerAct suitable for mobile manipulation, we tailored PerACt to predict goals in a moving camera frame rather than a static global frame. We also incorporate \method{}'s augmentations, including brightness, saturation, contrast, and hue augmentation. These are in addition to the SE(3) augmentations already present in PerAct.
We opt to use an input image resolution of $640 \times 480$, as opposed to \method{}'s $224 \times 224$.
Furthermore, to ensure a fair comparison with \method{}, We integrate prompts containing action primitives into PerAct.

As seen in Table \ref{tab:ablations}, we additionally evaluate several ablations of \method{}, including removing the forces from the low-level controller (\textit{w/o forces}), training without data augmentation (\textit{w/o augmentation}), using only RGB inputs without depth (\textit{w/o depth}), not pre-training on ImageNet 21k (\textit{w/o pre-training}), and predicting visual-force goals from RGBD data only, without a text prompt (\textit{w/o text cond.}).

\subsection{Test Set Results}

We collect a test set in unseen environments containing several unseen objects (Figure \ref{fig:experiment-envs}). The test set comprises a representative set of 60 keyframes from each task. We evaluate performance on the test set by calculating the average fingertip distance. \method{} outperforms the PerAct baseline and all ablations on our fingertip distance metric, demonstrating that our representation is capable of generating accurate kinematic goals in unseen situations.



\subsection{Real-World Results}

When evaluated on several real-world tasks in novel environments with unseen object instances, \method{} achieves a task success rate of 81\%. According to the results presented in Table \ref{tab:ablations}, we demonstrate the significance of force as a modality for successfully executing tactile tasks, such as picking up a paperclip. Our findings indicate that by using the ForceSight planner, which incorporates force-related objective information, the robot achieves improved performance in these tasks. Conversely, when the low-level controller ignores the force objective information, the robot's ability to successfully complete the task is compromised. For example, using only kinematic objectives tends to result in items falling due to the lack of grip force consideration. Similarly, neglecting applied forces often results in excessive force or failure to make proper contact with surfaces like tables and drawers, thus worsening grasp success rates.


\begin{table}
\centering
\scriptsize 
\resizebox{\columnwidth}{!}{
\begin{tabular}{|c|c||c|c|c|} \hline
    & Offline Goal Prediction & \multicolumn{3}{|c|}{Full System with Real Robot}\\ \hline
    & Avg &  & RMSE & RMSE\\
    & Fingertip & Task & Applied & Grip\\
    & Dist. (m) & Success &  Force (N) & Force (N)\\\hline
    \network{} (Ours) & \textbf{0.036} & \textbf{81\//100 (81\%)} &  \textbf{0.404} & 1.524 \\\hline
    Perceiver-Actor \cite{shridhar2023perceiver} & 0.058 & - & - & - \\\hline
    w/o forces & \textbf{0.036} & 10\//20 (50\%) &  - & - \\\hline
    w/o augmentation & 0.049 & 5\//20 (25\%) &  1.181 & 1.759 \\\hline
    w/o depth & 0.063 & 4\//20 (20\%) &  1.493 & 1.32 \\\hline
    w/o pre-training & 0.078 & 4\//20 (20\%) &  0.583 & 1.576 \\\hline
    w/o text cond. & 0.075 & 0\//20 (0\%) &  0.907 & \textbf{1.260}\\\hline
\end{tabular}
}
\vspace{2mm}
\caption{Ablations of ForceSight. Each ablation is tested with 2 trials for each of the 10 real-world tasks for a total of 20 trials per ablation. \jeremy{PerAct results are preliminary, I've implemented PerAct's translation/rotation augmentation and color jitter and am retraining}}
\label{tab:ablations}
\vspace{-8mm}
\end{table}


\youliang{summarized the 2 paragraphs below}

We also conducted an experiment in which we compared performance of \method{} with and without force goals using the same initial conditions. The success rate dropped from 90\% (18/20) to 45\% (9/20) when force goals were ignored. The controlled conditions enabled us to directly compare the failures. For example, the robot applied too much force to the underlying surface or grasped above the paperclip. It also applied too much force to the person's hand and failed to release the object during object handovers. It also tended to attempt to lift an object prior to applying sufficient grip force.



Additionally, omitting depth input during training made the robot prone to early grasping due to compromised depth perception. Moreover, data augmentation was found crucial for better performance in varied environments and with unfamiliar objects.


\begin{figure}
  \centering
  \includegraphics[width=1\linewidth]{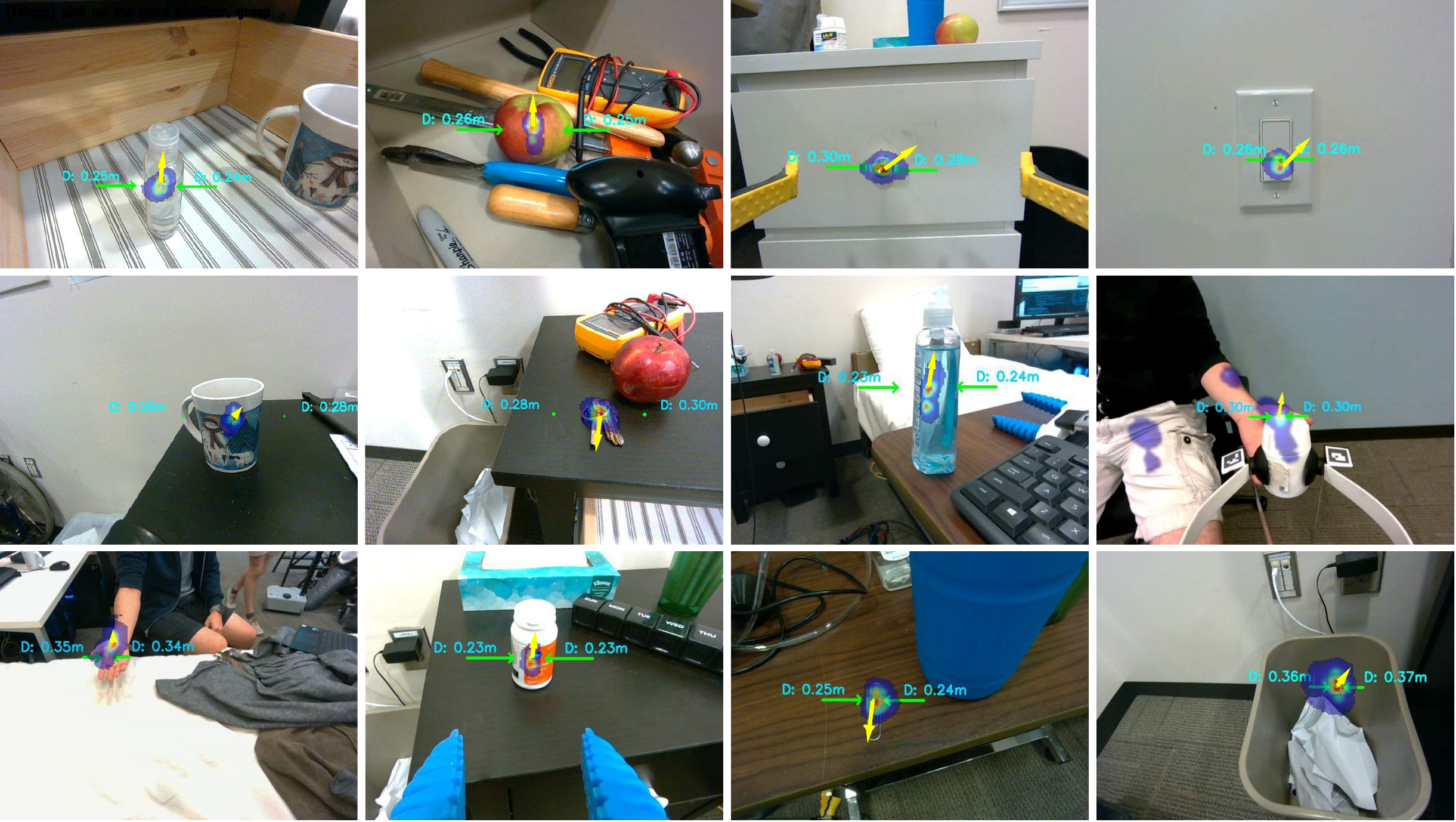}
  \vspace{-3mm}
  \caption{\network{} is capable of providing accurate estimations of fingertip location, applied forces, and grasping forces associated with a task in a variety of environments. Our network additionally generalizes to unseen object instances, does not depend on a specific camera setup, and works in situations with partial occlusion. \jeremy{add prompts?}}
  \label{fig:All_tasks}
\vspace{-5mm}
\end{figure}


\section{Limitations}

While \network{} exhibits encouraging performance across diverse and challenging tasks, we recognize certain limitations and areas for future improvement.

One recurrent failure mode observed in our model pertains to inaccuracies in our depth predictions, particularly those further away from the camera (1m). Despite that, the adaptive capability of visual servoing largely accounts for these discrepancies, as errors diminish at closer distances.

Another limitation of the current \method{} model is the requirement for targets to be within the camera's field of view, which limits performance on some tasks. This constraint may limit task performance in scenarios such as drawer opening, where predictions are sometimes clipped to the image's edge.


Our keyframe representations do not provide complete information about the gripper's pose, assuming pitch and roll to be constant. However, this can be addressed by adding additional MLP heads to the model's output. 



\youliang{summarized 2 paragraphs}

While our study showcases success in real-world tasks like pick-and-place, it remains limited in scope. However, the scalability of transformer models and our efficient data collection suggest the potential for broader complex tasks. Additionally, our experiments, conducted solely with the RE1 robot, should be expanded to different robotic platforms in future studies to truly gauge our model's versatility and robustness.


\section{Discussion}
\begin{figure}[H]
  \centering
  \includegraphics[width=1\linewidth]{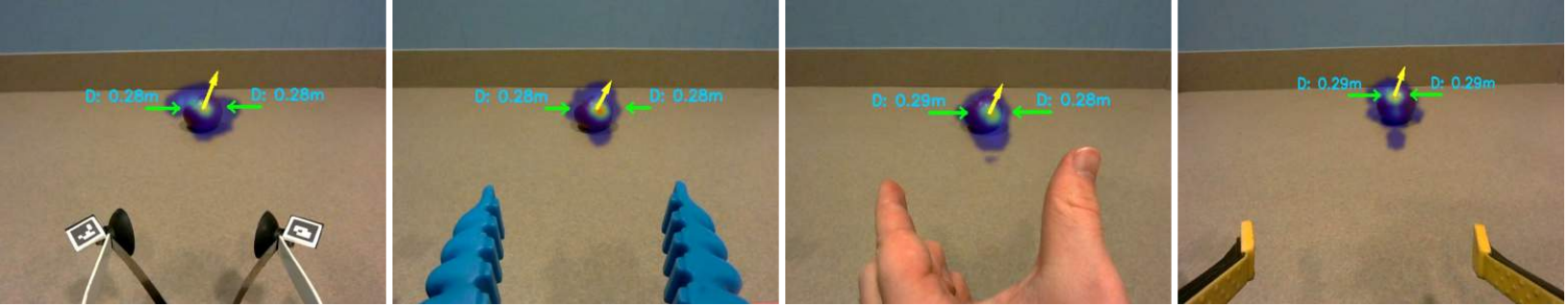}
  \vspace{-5mm}
  \caption{Predictions from \method{} are agnostic to the agent and camera perspective, as shown in this example for the apple grasping task.}
  \label{fig:agent-agnostic}
\end{figure}

\begin{figure}[H]
  \centering
  \includegraphics[width=.75\linewidth]{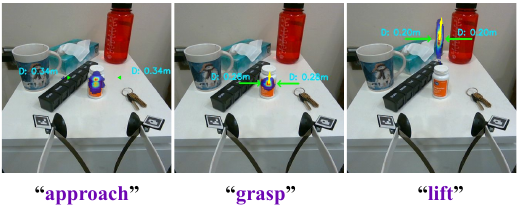}
  \vspace{-2mm}
  \caption{We observe that \method{} is able to make predictions for action primitives that are more than one keyframe into the future, despite having been trained to predict goals associated with the next keyframe.}
  \label{fig:multistep-pred}
   \vspace{-2mm}
\end{figure}

After training, we observed multiple emergent behaviors exhibited by \network{} (Figure \ref{fig:agent-agnostic} and \ref{fig:multistep-pred}). Most notable was \method{}'s ability to generalize to unseen object instances despite only having observed one type of each object during training, likely due to the high-level visual associations retained from pre-training. \network{} also exhibited adaptability in its treatment of action primitives in relation to various objects. \network{} managed to apply action primitives to objects without any explicit representation of such actions during the training phase. For example, if given the action primitive ``grasp" in combination with a prompt concerning a light switch or the action primitive ``push" in combination with a prompt concerning a medicine bottle, the model responded with predictions corresponding to the specified actions, despite these specific subtasks being absent in the training data.

\section{Conclusion}
\label{sec:conclusion}
We presented \method{}, a text-conditioned robotic planner that generates tactile and kinematic goals to enable the execution of multiple contact-rich tasks, generalizing to unseen environments and new object instances. We demonstrated the usefulness of \method{} with a series of real-world robotic tasks, and show that the use of tactile objectives improves performance on these tasks.

\clearpage


\clearpage

\bibliographystyle{IEEEtran}
\bibliography{cited}

\end{document}